# Improving Entity Recognition Using Ensembles of Deep Learning and Fine-tuned Large Language Models: A Case Study on Adverse Event Extraction from Multiple Sources

Yiming Li, MS, [1] Deepthi Viswaroopan, MD, [1] William He, [2] Jianfu Li, PhD [3], Xu Zuo, PhD, [1] Hua Xu, PhD, [4] Cui Tao, PhD,[3]*

*corresponding author

[1]McWilliams School of Biomedical Informatics, The University of Texas Health Science Center at Houston, Houston, TX 77030, USA

[2]Department of Electrical & Computer Engineering, Pratt School of Engineering, Duke University, 305 Tower Engineering Building, Durham, NC 27708, USA

[3]Department of Artificial Intelligence and Informatics, Mayo Clinic, Jacksonville, FL 32224, USA

[4]Section of Biomedical Informatics and Data Science, School of Medicine, Yale University, New Haven, CT 06510, USA

**Corresponding Author:**

Cui Tao, PhD, Department of Artificial Intelligence and Informatics, Mayo Clinic, 4500 San Pablo Rd, Jacksonville, FL 32224, USA; e-mail: Tao.Cui@mayo.edu

# ABSTRACT

**Objective**

Adverse event (AE) extraction following COVID-19 vaccines from text data is crucial for monitoring and analyzing the safety profiles of immunizations, identifying potential risks and ensuring the safe use of these products. Traditional deep learning models are adept at learning intricate feature representations and dependencies in sequential data, but often require extensive labeled data. In contrast, large language models (LLMs) excel in understanding contextual information, but exhibit unstable performance on named entity recognition (NER) tasks, possibly due to their broad but unspecific training. This study aims to evaluate the effectiveness of LLMs and traditional deep learning models in AE extraction, and to assess the impact of ensembling these models on performance.

**Methods**

In this study, we utilized reports and posts from the Vaccine Adverse Event Reporting System (VAERS) (n=621), Twitter (n=9,133), and Reddit (n=131) as our corpora. Our goal was to extract three types of entities: vaccine, shot, and adverse event (ae). We explored and fine-tuned (except GPT-4) multiple LLMs, including GPT-2, GPT-3.5, GPT-4, Llama-2 7b, and Llama-2 13b, as well as traditional deep learning models like Recurrent neural network (RNN) and Bidirectional Encoder Representations from Transformers for Biomedical Text Mining (BioBERT). To enhance performance, we created ensembles of the three models with the best performance. For evaluation, we used strict and relaxed F1 scores to evaluate the performance for each entity type, and micro-average F1 was used to assess the overall performance.

**Results**

The ensemble model achieved the highest performance in "vaccine," "shot," and "ae," with strict F1-scores of 0.878, 0.930, and 0.925, respectively, along with a micro-average score of 0.903. These results underscore the significance of fine-tuning models for specific tasks and demonstrate the effectiveness of ensemble methods in enhancing performance.

**Conclusion**

In conclusion, this study demonstrates the effectiveness and robustness of ensembling fine-tuned traditional deep learning models and LLMs, for extracting AE-related information following COVID-19 vaccination. This study contributes to the advancement of natural language processing in the biomedical domain, providing valuable insights into improving AE extraction from text data for pharmacovigilance and public health surveillance.

# INTRODUCTION

The COVID-19 pandemic has posed a significant global health threat, with over 111.8 million confirmed cases and more than 1.1 million deaths reported in the United States as of April, 2024 [1], [2]. The severity of COVID-19 is evident from its diverse symptoms, including fever, shortness of breath, fatigue, body aches, loss of taste or smell, sore throat, nausea, and diarrhea, which can progress to severe respiratory distress syndrome and death [3, p. 19], [4]. As with other infectious diseases, vaccination has emerged as the most effective measure to control the spread of COVID-19 and reduce its impact on public health [5], [6], [7], [8], [9], [10]. In the United States, by May 2023, over 676 million doses of COVID-19 vaccines had been administered, with approximately 81.4% of the population receiving at least one dose and 69.5% fully vaccinated [2]. However, the introduction of COVID-19 vaccines has been accompanied by reports of adverse events (AEs). As of December 2022, the Vaccine Adverse Event Reporting System (VAERS) in the United States has received over 900 thousand reports of AEs following COVID-19 vaccination [7], [8], [9]. Of them, 4.7% were classified as serious reports [7]. Common AEs following COVID-19 vaccination include mild symptoms such as fever, fatigue, and injection site pain [11]. However, there have been reports of severe AEs, including anaphylaxis and myocarditis, though these are rare [12]. Understanding the AEs following COVID-19 vaccination is crucial for ensuring the safety and efficacy of vaccination campaigns [13].

For AE following immunization reporting, VAERS is a crucial surveillance program managed by the Centers for Disease Control and Prevention (CDC) and the U.S. Food and Drug Administration (FDA) [14]. It serves as a cornerstone for monitoring the safety of vaccines licensed in the United States [15]. VAERS collects and analyzes reports of adverse events (possible side effects or health problems) that occur after vaccination [16]. Healthcare providers, vaccine manufacturers, and the public can submit reports to VAERS, which is critical for identifying potential safety concerns and ensuring the ongoing safety of vaccines [17, pp. 2000–2013].

Nowadays, social media has become a common platform for people to exchange ideas and express feelings [18]. With the COVID-19 pandemic, there has been a significant increase in the number of individuals using social media to share their experiences related to COVID-19, including symptoms following vaccination [19], [20]. Posting on social media is often less trivial than reporting to formal systems like VAERS, making it a valuable source of real-time information on vaccine safety and AEs. During the outbreak, there have been over 468+ million million posts related to COVID-19 on various social media platforms, highlighting the widespread use of social media for discussing pandemic-related topics [21]. Understanding the content and sentiment of these posts can provide additional insights into the public's perception and concerns regarding COVID-19 vaccines, complementing traditional surveillance systems [22].

The recent advancements in natural language processing (NLP) have led to the development of powerful large language models (LLMs) such as the Generative Pre-trained Transformer (GPT) series. These models, trained on vast amounts of text data, have shown remarkable capabilities in understanding and generating human-like text [23], [24], [25]. Li et al. utilized LLMs to extract the relations for acupuncture point locations with a high performance [26]. Wang et al. proposed GPT-NER to improve Named Entity Recognition (NER) performance using LLM [27]. GPT-NER transforms the sequence labeling task of NER into a text-generation task, allowing LLMs to adapt more easily [27].

Additionally, GPT models can be fine-tuned to specific tasks, such as extracting AEs, by providing them with labeled examples of the information to extract [28]. For instance, researchers can annotate social media posts to indicate whether they contain information about adverse events following COVID-19 vaccination. Li et al. investigated multiple pre-trained and fine-tuned LLMs to extract the AE-related information in the VAERS reports, resulting AE-GPT achieved 0.816 for relaxed match [28]. By training a GPT model on these annotated examples, it can learn to identify and extract relevant information from new, unseen reports. Given the large volume of social media posts related to COVID-19, there is a growing interest in exploring the use of GPT models to extract information, including AEs, from these text data.

Using GPT for AE extraction from diverse sources offers several advantages. Firstly, GPT can process a large volume of text data quickly, allowing for the analysis of a vast number of social media posts in a short period [29]. Secondly, GPT's ability to understand context and generate human-like text enables it to capture nuanced information, such as the severity of AEs or the context in which they occurred [30]. Lastly, GPT can be continuously updated and improved as new data becomes available, ensuring that it remains effective in extracting AEs from evolving social media discussions [31]. However, the performance of LLMs on NER tasks has shown considerable variability, presenting challenges in consistently achieving high accuracy. For example, Hu et al. explored ChatGPT's potential for clinical named entity recognition in the 2010 i2b2 challenge using two prompt strategies [32]. They compared its performance to BioClinicalBERT using synthetic clinical notes and found ChatGPT's performance was lower (F1 scores of 0.620 vs. 0.888) [32]. Monteiro and Zanchettin explored approaches to enhance NER in transformer-based models pre-trained for language modeling [33]. Despite the remarkable reasoning abilities demonstrated by GPT 3.5 and "ChatGPT" models, their quantitative performance was found to still lag behind that of traditionally fine-tuned models in their task [33]. This instability is particularly problematic in the biomedical domain, where precise identification of entities such as drug names, diseases, and patient attributes is critical for effective data analysis and decision-making.

Given the challenges, in this study, we will firstly employ LLMs as well as traditional models to conduct the task of identifying entities related to AEs, such as *vaccines*, *shots*, and *adverse events*, from a large corpus of text. Additionally, we will also explore whether an ensemble approach, combining the strengths of LLMs with deep learning models, can enhance the stability and overall performance of the NER task. This research aims to contribute to the growing body of literature on utilizing advanced NLP techniques for pharmacovigilance and vaccine safety monitoring.

## METHODS

In this study, we selected the VAERS and social media as data sources. We annotated the vaccine-related entities and predicted them using both pretrained and finetuned traditional deep learning models, as well as large language models. Figure 1 provides an overview of the framework.

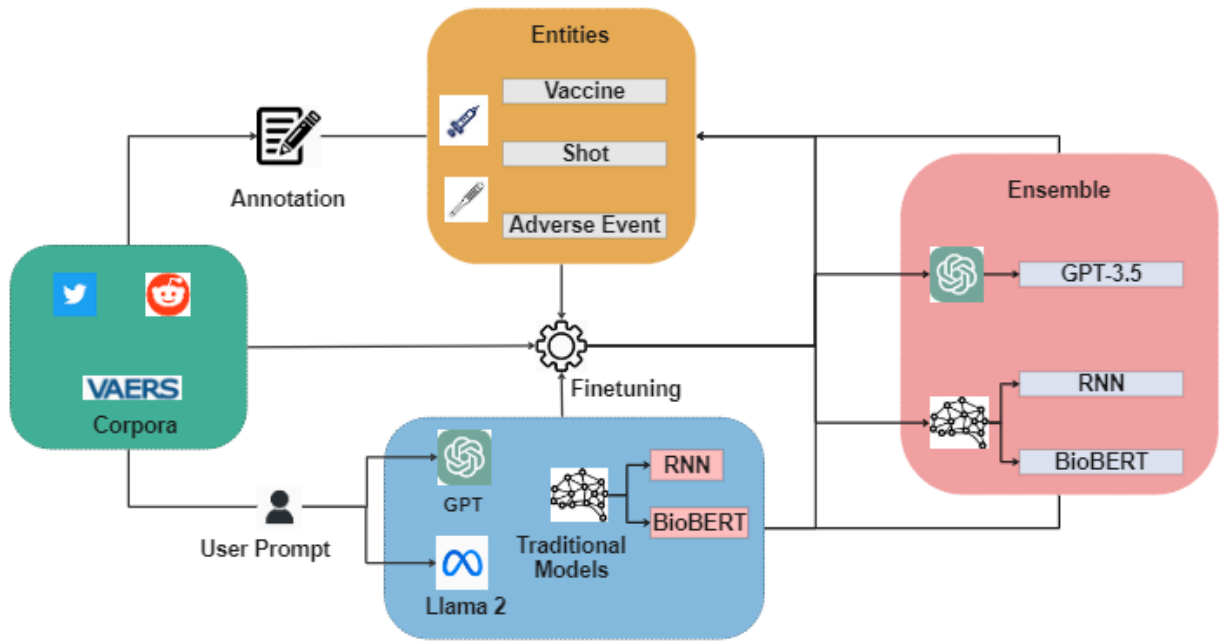

Figure 1 Overview of the framework

## Data sources

In this study, we utilized the reports from VAERS, and posts from Twitter and Reddit as our primary data sources.

### VAERS

VAERS is a national vaccine safety surveillance program that collects and analyzes information about adverse events following immunization (AEFI). VAERS reports consist of three Comma-Separated-Value (CSV) files - VAERSDATA.CSV, VAERSVAX.CSV, and VAERSSYMPTOMS.CSV - grouped by year. The VAERSDATA.CSV contains demographic information, vaccination and AE timing, symptom descriptions, allergy history, and serious outcomes. VAERSVAX.CSV provides details on vaccine type and manufacturer for each adverse event, while VAERSSYMPTOMS.CSV lists the symptoms associated with each AEFI, as mapped from the Preferred Term (PT) in the MedDRA terminology. The three tables are linked by the primary key 'VAERS_ID'.

We utilized the dataset from the study conducted by Li et al., which comprised VAERS reports related to COVID-19 vaccination (VAX_TYPE="COVID19" in "VAERSVAX.CSV") collected from December 13, 2020, to December 28, 2022, amounting to 900,522 reports [7]. From these selected datasets, we randomly selected 230 VAERS reports and extracted the SYMPTOM_TEXT field in VAERSDATA.CSV for subsequent text analysis.

## Twitter

Twitter, a widely-used microblogging platform, has become an invaluable source for real-time information and public sentiment analysis [34]. The platform sees around 500 million tweets daily, covering a wide array of topics, including health-related discussions and public health issues [35]. As of March 2019, Twitter boasts approximately 330 million monthly active users globally [36]. Its vast user base and capacity to swiftly disseminate information make Twitter an essential resource for analyzing public health trends, including adverse events following COVID-19 vaccination.

In this study, we reused the data collected by Lian et al. [37] They utilized the Twitter streaming API to gather posts related to vaccines.

The inclusion criteria for our data collection were as follows:
- Tweets from December 2020 to August 2021;
- Include a predefined set of keywords:  Pfizer, Moderna, J&J, Johnson & Johnson, BioNTech, vaccine, AstraZeneca, covidvac, etc.

Due to the high noise level in the Twitter data, which included tweets unrelated to AEs following COVID-19 vaccine, exclusion criteria were applied to identify content specifically related to personal experiences of post-COVID-19 vaccine AEs.

- Retweets and quotes were removed.
- Tweets from users with over 10,000 followers, classified by Twitter as "super follows," were excluded.

Following the screening process, a total of 111,229 tweets were retained for analysis.

## Reddit

In addition to Twitter, we also utilized Reddit streaming API to collect posts from Reddit for AE extraction. Reddit, a popular social news aggregation, web content rating, and discussion website, features a vast user base and sees a significant flow of posts daily [38, p. 19]. With approximately 52 million daily active users and numerous topic-specific communities known as "subreddits," Reddit offers a diverse range of discussions, including those related to health and COVID-19 vaccines [39], [40].

We collected data from Reddit posts within the time frame starting from December 1, 2020, to December 31, 2022. To include qualified Reddit posts in our study, we focused on all subreddits, which encompasses a wide range of discussions and is not limited to a particular topic. This approach ensured that we captured a broad spectrum of posts related to COVID-19 vaccines. Additionally, we utilized three sets of keywords to filter posts: keywords related to COVID-19 (e.g., "COVID19," "COVID," "Covid-19") to target posts discussing the pandemic and its vaccines, keywords pertaining to vaccine manufacturers (e.g., "Moderna," "Pfizer," "Johnson," "Janssen," "AstraZeneca,"

"Novavax") to identify posts discussing these specific vaccines, and self-related keywords (e.g., "I," "my," "mine," "me," "myself") to identify posts where users shared their personal experiences or opinions regarding COVID-19 vaccines.

Finally, we retained and annotated randomly the posts and reports that contained at least one AE keyword related to the COVID-19 vaccine. This dataset includes 621 reports from VAERS, 9,133 tweets, and 131 posts from Reddit.

## Annotation

In this study, we employed CLAMP (Clinical Language Annotation, Modeling, and Processing) to annotate COVID-19 vaccine-related AEs for corpora [41]. These named entities included *vaccine*, *shot*, and *ae* (adverse event). The definition and examples of entities are shown in Table 1. The *vaccine* entity referred to the specific COVID-19 vaccine mentioned in the posts/reports, with the full name of the vaccine selected if possible. Examples of *vaccine* entities included "Pfizer vaccine," "Moderna vaccine," among others. The *shot* entity indicated which shot of the COVID-19 vaccine was being referred to in the posts/reports, such as "first dose," "second dose," and so on. *ae* entities denoted the symptoms or diseases experienced following vaccination, with annotations including "fever," "sore arm," "headache," and similar symptoms.

During the annotation process, we adhered to specific guidelines. These guidelines instructed annotators not to include space and special characters in the text, avoid annotating adjectives for a concept (e.g., mild, strong), and annotate only those AEs that were the symptoms or diseases experienced by the vaccine recipient, excluding those experienced by family members or others.

An annotation example related to adverse events following COVID-19 vaccination is provided in Figure 2.

Table 1 Definition and examples of entities

| Entity | Definition | Examples |
| --- | --- | --- |
| **vaccine** | the specific COVID-19 vaccine mentioned in the posts/reports | Moderna vaccine, Pfizer vaccine, Coronavirus Vaccine |
| **shot** | Specific dose(s) of a vaccine administered through an injection | booster shot, 1st dose, 2nd dose |
| **ae** | Any negative or unexpected medical occurrence or side effect that follows vaccination. | sore arm, headache, fever |

My **first shot** of **moderna vaccine** = very **sore arm** the next day and very **tired** so **I slept a lot**.
 • dose          • vaccine          • ae                                • ae        • ae

These are the known side effects that I've also read and heard from my coworkers . In our line of work , its more important to protect the people we work with .

Figure 2 An annotation example in this study

## Models

### GPT

The GPT model, developed by OpenAI, has demonstrated remarkable capabilities in various NLP tasks, including NER [28], [32]. GPT, based on the Transformer architecture, leverages its extensive pre-training strategy to learn contextual representations of words, enabling it to understand the context in which named entities appear [42], [43]. This contextual understanding allows GPT to effectively identify and classify named entities in text, making it a powerful tool for NER tasks in NLP.

## Llama-2

Llama-2, developed by Microsoft, offers models with 7 billion, 13 billion, and 70 billion parameters, providing powerful capabilities across various NLP tasks [28]. This range of models allows Llama to excel in tasks such as text classification, language understanding, and text generation [44].

## RNN

RNNs are designed to process sequential data by maintaining an internal memory [45]. This memory allows them to learn patterns and dependencies in sequences, making them suitable for tasks like speech recognition, language translation, and time series prediction [46]. RNNs process data step by step, updating their internal state with each new input [47]. This ability to remember past information makes them effective in tasks where context is important. However, they can struggle with long sequences due to the vanishing gradient problem, which limits their ability to retain information over long periods [47].

## BioBERT

BioBERT is a variant of the BERT (Bidirectional Encoder Representations from Transformers) model that has been specifically pre-trained on biomedical text [48]. This pre-training helps BioBERT better understand the complex language used in biomedical literature, making it particularly effective for tasks in the biomedical domain [49]. By fine-tuning BioBERT on specific tasks or datasets, researchers can leverage its biomedical knowledge to achieve state-of-the-art results in various NLP tasks, such as NER, relation extraction, and question answering in the biomedical field [48]. BioBERT's ability to capture domain-specific nuances and terminology makes it a valuable tool for advancing research and applications in biomedicine and healthcare.

**Experiment setup**

## Data Split

In this study, we split the dataset into training, validation, and test sets using an 8:1:1 ratio. Figure 3 provides a detailed breakdown of the number of entities for each set.

## GPT

For the GPT models, we employed pre-trained versions of GPT-2, GPT-3.5, and GPT-4. Additionally, we fine-tuned the pre-trained GPT-2 and GPT-3.5 models for our specific task.

For the prompts, we divided them into two styles: split and merged. In the split style, prompts were designed to extract entities individually, focusing on one entity at a time. In contrast, the merged style involved prompts that aimed to extract all entities simultaneously. We determined the best-performing prompt style for each GPT model based on experimental results, as detailed in Supplementary Table S1. This approach allowed us to tailor the prompt style to each model's strengths, optimizing performance for this entity extraction tasks.

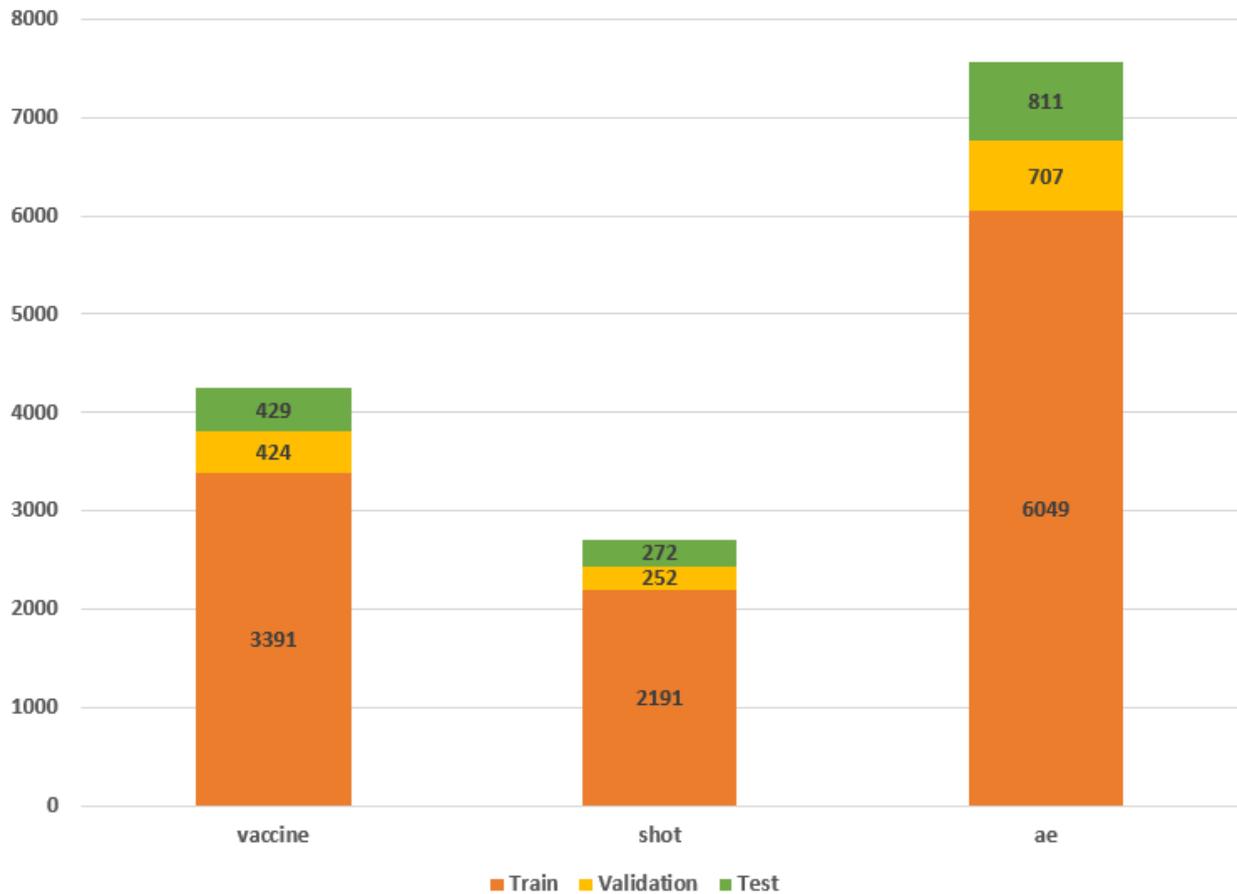

Figure 3 A breakdown of the number of entities for each set

Split prompt style was structured as follows, with the *vaccine* entity for fine-tuned GPT-2 used as an example:

> "question": "Please extract all the names of vaccine from the following note"
>
> "context": {note}

Merged prompt style (pretrained GPT 2) was formatted as follows:

> *"Please extract all names of dose, vaccine, and adverse event from this note, and put them in a list: {note}"*

In our experiment setup, we used several important parameters for generating text with the GPT models. The temperature parameter, controls the randomness of the generated output. A lower temperature results in more deterministic outputs, while a higher temperature leads to more diverse but potentially less coherent text. The temperature parameter allows us to balance between generating text that closely resembles the training data and generating more novel, creative outputs. We also utilized the maximum output token parameter, which defines the maximum length of the generated text in terms of the number of tokens. A token in this context refers to the smallest unit of text that the model can process, which can be a word, subword, or character depending on the tokenization scheme used. Setting a maximum output token limit helps control the length of the generated text, ensuring that it remains within a manageable length for readability and analysis.These settings were selected based on preliminary experiments to achieve a balance between text quality and computational efficiency. Supplementary Table S2 shows the important parameters used by each GPT model.

## Llama 2

For Llama 2, we utilized two variants: Llama 2 7b and Llama 2 13b, both of which were fine-tuned for our task. The temperature setting for all Llama 2 models was set to 1. The specific prompts used are shown in Supplementary Table S1. During the inference phase, we employed prompts in split mode to extract entities. For the fine-tuning process of Llama 2 to extract *vaccine*, we used the following format:

> "instruction": "Please extract all names of vaccines",
>
> "input": {note}.

For the remaining parameters used in Llama 2, please refer to Supplementary Table S3.

## RNN

We fine-tuned a Recurrent Neural Network (RNN) for our experiments with the following configuration: lowercase words were set to 1, replacing digits with 0 was enabled (zeros=1), and the character embedding dimension was set to 25 (char_dim=25). The character LSTM hidden layer size was also set to 25 (char_lstm_dim=25), and a bidirectional LSTM for characters was used (char_bidirect=1). For token embeddings, the word embedding dimension was 200 (word_dim=200), and the token LSTM hidden layer size was 100 (word_lstm_dim=100). A bidirectional LSTM for words was employed (word_bidirect=1), and a Conditional Random Field (CRF) was used for tagging (crf=1). Dropout with a rate of 0.5 was applied (dropout='0.5'). The tagging scheme used was IOB (tag_scheme='iob'), and the model was fine-tuned over 30 epochs (epoch=30).

## BioBERT

We also fine-tuned BioBERT v1.1 for our experiments. The configuration used for fine-tuning BioBERT v1.1 included an attention dropout probability of 0.1, a hidden activation function of "gelu," a hidden dropout probability of 0.1, a hidden size of 768, an initializer range of 0.02, an intermediate size of 3,072, a maximum position embeddings of 512, 12 attention heads, 12 hidden layers, a type vocabulary size of 2, and a vocabulary size of 28,996.

## Ensemble

Ensembling enables us to capitalize on the strengths of each individual model, enhancing overall performance through the amalgamation of their predictions. In this study, we utilized ensembles comprising fine-tuned GPT-3.5, RNN, and BioBERT models, employing a majority voting scheme. This approach entails consolidating the predictions from each model and selecting the most frequently predicted outcome as the final result.

The experiments for the GPT models and BioBERT were conducted on a server with 8 Nvidia A100 GPUs, each offering 80GB of memory. Meanwhile, the Llama models and RNN were executed on a server equipped with 5 Nvidia V100 GPUs, each providing a memory capacity of 32GB.

**Evaluation**

In this study, we employed inter-rater agreement to measure the consistency between annotators in the annotation process. This metric helps assess the reliability of the annotations and ensures that the data is accurately labeled for further analysis. To evaluate the performance of the models in entity recognition, we used F1 scores, calculated both in relaxed and strict settings, for each entity type. The relaxed F1 score allows for some leniency in matching predicted and ground truth entities, while the strict F1 score requires an exact match. Additionally, we calculated the micro-average F1 score to assess the overall performance of each model across all entities. These evaluation metrics provide a comprehensive understanding of how well each model performs in recognizing adverse event-related entities following COVID-19 vaccination.

## RESULTS

Table 2 shows the inter-rater agreement for the annotation of entities. The agreement for the entity *vaccine* was 0.94, indicating strong agreement between annotators. For the entity *shot*, the agreement was perfect, with a score of 1, indicating complete agreement. The agreement for the entity *ae* was slightly lower at 0.73, indicating moderate agreement. Overall, the inter-rater agreement across all entities was 0.77, reflecting a substantial level of agreement between annotators.

Table 2 Inter-rater agreement

| Entities | Inter-rater agreement |
|---|---|
| vaccine | 0.94 |
| shot | 1 |
| ae | 0.73 |
| Overall | 0.77 |

Table 3 (a) presents the relaxed F1 of different models in extracting AE-related information following COVID-19 vaccines. BioBERT attained the highest F1-score for the keyword *vaccine* at 0.925, while ensemble exhibited the best performance for the keyword *shot* with an F1-score of 0.965, and for *ae* with a score of 0.934. Overall, ensembles showed the highest micro-average F1-score of 0.926, underscoring their effectiveness in extracting AE information from social media posts related to COVID-19 vaccines.

Table 3(b) shows the strict F1 scores for various models across different entities, including the overall micro-average F1 score. The comparison highlights the performance differences between pretrained and fine-tuned models. In general, fine-tuned models tend to outperform their pretrained counterparts. Ensemble achieved the highest performance in *vaccine*, *shot*, *ae*, and micro-average with 0.878, 0.930, 0.925, and 0.903 respectively.

Table 3(a) Relaxed F1 of different models in extracting AE-related information following COVID-19 vaccines.

|  | Pretrained GPT-2 | Finetuned GPT-2 | Pretrained GPT-3.5 | Fine-tuned GPT-3.5 | Pretrained GPT-4 | Pretrained Llama 2 7b | Finetuned Llama 2 7b | Pretrained Llama 2 13b | Fine-tuned Llama 2 13b | RNN | BioBERT | Ensemble |
| --- | --- | --- | --- | --- | --- | --- | --- | --- | --- | --- | --- | --- |
| vaccine | 0 | 0.827 | 0.446 | 0.849 | 0.492 | 0.319 | 0.335 | 0.524 | 0.739 | 0.896 | 0.925 | 0.918 |
| shot | 0 | 0.868 | 0.416 | 0.917 | 0.441 | 0.008 | 0.156 | 0.117 | 0.176 | 0.917 | 0.936 | 0.965 |
| ae | 0 | 0.512 | 0.400 | 0.648 | 0.417 | 0.048 | 0.368 | 0.150 | 0.348 | 0.871 | 0.905 | 0.934 |
| Micro-average | 0 | 0.687 | 0.412 | 0.758 | 0.437 | 0.134 | 0.238 | 0.252 | 0.398 | 0.886 | 0.916 | 0.926 |

Table 3(b) Strict F1 of different models in extracting AE-related information following COVID-19 vaccines.

|  | Pretrained GPT-2 | Finetuned GPT-2 | Pretrained GPT-3.5 | Fine-tuned GPT-3.5 | Pretrained GPT-4 | Pretrained Llama 2 7b | Finetuned Llama 2 7b | Pretrained Llama2 13b | Fine-tuned Llama2 13b | RNN | BioBERT | Ensemble |
| --- | --- | --- | --- | --- | --- | --- | --- | --- | --- | --- | --- | --- |

| | | | | | | | | | | | |
|---|---|---|---|---|---|---|---|---|---|---|---|
| vaccine | 0 | 0.723 | 0.182 | 0.781 | 0.212 | 0.172 | 0.187 | 0.359 | 0.536 | 0.847 | 0.864 | 0.878 |
| shot | 0 | 0.766 | 0.359 | 0.853 | 0.360 | 0 | 0.094 | 0.075 | 0.137 | 0.860 | 0.875 | 0.930 |
| ae | 0 | 0.355 | 0.269 | 0.632 | 0.270 | 0.034 | 0.288 | 0.103 | 0.274 | 0.763 | 0.786 | 0.925 |
| Micro-average | 0 | 0.559 | 0.258 | 0.716 | 0.263 | 0.082 | 0.163 | 0.176 | 0.296 | 0.805 | 0.824 | 0.903 |

# DISCUSSION

## Findings

In this study, we explored the use of both state-of-the-art traditional deep learning models and large language models for extracting AE-related information following COVID-19 vaccination from social media posts. Our findings highlight several key points regarding the performance and effectiveness of these models in this NER task.

Fine-tuning of pre-trained LLMs, such as GPT-2 and GPT-3.5, played a pivotal role in enhancing their ability to recognize entities related to AEs. This process allows the models to adapt to the specific characteristics of the dataset, leading to improved performance. This finding is consistent with prior work by Li et al., who previously explored pretrained and fine-tuned LLMs for extracting adverse events following influenza vaccines [28]. Interestingly, we observed minimal differences in performance among different grades of GPT models, suggesting a certain level of robustness in handling NER tasks regardless of the model's size.

In contrast, Llama models exhibited more noticeable differences in performance, which can be attributed to their specialized architecture and training objectives for medical NLP tasks. This highlights the importance of selecting Llama 2 models tailored to the specific task at hand, as the performance may vary based on the model's design and training data.

In this study, we also investigated the effectiveness of ensembling fine-tuned LLMs with traditional deep learning models for the NER task related to AEs following COVID-19 vaccination from social media posts. Our findings reveal the significant implications of ensembling in improving the overall performance of the models. While LLMs have shown to be inferior to traditional deep learning models in this NER task, ensembling the two types of models led to a substantial improvement in the strict F1 score. Specifically, the ensembling approach resulted in an 8% increase in the strict F1 score, exceeding 90%. This improvement underscores the complementary nature of LLMs and traditional deep learning models, suggesting that combining their strengths can lead to better performance than either model alone. The effectiveness of ensembling can be attributed to the unique strengths of each type of model. LLMs excel in capturing complex linguistic patterns and contextual information, making them highly effective in understanding the nuances of social media posts. On the other hand, traditional deep learning models, with their robust architectures and ability to learn complex feature representations, complement LLMs by providing additional context and generalization capabilities. Furthermore, ensembling helps mitigate the weaknesses of individual models. While LLMs may struggle with certain aspects of the NER task, such as entity ambiguity or rare occurrences, traditional deep learning models can compensate for these limitations by providing more robust and reliable predictions in such cases [50], [51]. This complementary nature of the two types of models makes ensembling a powerful strategy for improving overall performance.

Despite the overall success in NER, we identified two entities, *shot* and *ae*, that did not perform as well as expected. The *shot* entity, although showing high inter-rater agreement, was rare in occurrence, potentially affecting the models' ability to learn from sufficient examples. Similarly, the *ae* entity's broad and ambiguous nature posed challenges for accurate recognition, leading to lower performance in this category.

## Error analysis

The error analysis (Table 4) shows that the model struggled most with false negatives for *ae*, missing 9.12% of human-annotated entities. For *vaccine* and *shot*, false positives were more prevalent, with 6.56% and 11.43%, respectively, of machine-annotated entities being incorrect. Boundary mismatches were relatively low across all entities, indicating that the model generally identified entity boundaries accurately. No instances of incorrect entity types were identified for any entity, indicating that when the model made a prediction, it tended to assign the correct entity type.

In our ensemble method, we achieved near-perfect performance for AE extraction. However, there were instances where the model missed certain AE-related entities, such as "flu-like symptoms", "cold", "shaky", "achiness", and "a knot in my hairline". These entities may not be explicitly defined in an AE-related ontology, leading to their neglect. Additionally, the model occasionally misclassified the general term "vaccine" as a specific entity type *vaccine*.

These errors can be attributed to several factors. The neglect of certain entities may be due to the limited scope of the model's training data, which may not have included these specific entities. The misclassification of "vaccine" may be caused by the ambiguity of the term, which can refer to both the general concept of vaccination and specific instances of vaccines.

To address these errors, expanding the training data to include a more diverse range of AE-related entities could improve the model's performance. Additionally, refining the model's entity recognition capabilities to better distinguish between general terms and specific entities, such as using context-aware techniques, could help reduce misclassifications.

Table 4 Error analysis by each entity type

|  | Boundary Mismatch (out of human annotated entities) | False Positive (out of machine annotated entities) | False Negative (out of human annotated entities) | Incorrect Entity Type (out of machine annotated entities) |
| --- | --- | --- | --- | --- |
| *vaccine* | 23/429, 5.36% | 29/442, 6.56% | 26/429, 6.06% | 0/442, 0% |
| *shot* | 16/272, 5.88% | 36/315, 11.43% | 6/272, 2.21% | 0/315, 0% |
| *ae* | 37/811, 4.56% | 41/792, 5.18% | 74/811, 9.12% | 0/792, 0% |

## Strengths and limitations

This study has several strengths that contribute to its significance in the field of AE detection from VAERS and social media data. Firstly, the utilization of three years of patient self-reported data, which includes reports from VAERS and two social media platforms, adds a robust and comprehensive dimension to the analysis. This extensive dataset allows for a thorough examination of AE reports over time, providing valuable insights into trends and patterns that may not be apparent in shorter-term studies. Additionally, the inclusion of social media data alongside VAERS reports offers a more holistic view of AEs, capturing a wider range of patient experiences and opinions.

Secondly, this study stands out for its comprehensive exploration of a wide range of traditional deep learning and LLMs. By examining and comparing the performance of these models, the study sheds light on their respective strengths and weaknesses in the context of NER tasks related to AEs. This comprehensive analysis not only provides valuable insights for researchers and practitioners in the field but also serves as a reference point for future studies looking to employ similar models. Overall, these strengths underscore the thoroughness and rigor of the study, enhancing its credibility and relevance in the field.

The study has several limitations that should be considered. Firstly, the corpora used for AE extraction may be incomplete or contain inaccuracies, leading to misclassification or omission of relevant events. The reliance on social media data introduces bias, as users may not be representative and reporting may be influenced by various factors. Moreover, while the study demonstrates the feasibility of using LLMs for AE extraction in the context of COVID-19 vaccines, its generalizability to other medical domains may be limited and require further validation.

# CONCLUSION

In conclusion, this study contributes by exploring the effectiveness of both LLMs and traditional deep learning models in the AE extraction task. It also highlights the significant improvement achieved by ensembling fine-tuned LLMs and traditional deep learning models. The findings provide valuable insights for both biomedical informatics and clinics, suggesting that ensembles of these models can significantly enhance the accuracy and robustness of AE extraction from text data, thereby supporting clinical decision-making and pharmacovigilance efforts. Future work should focus on further refining the LLM models and expanding the dataset to enhance the performance and generalizability.

# ACKNOWLEDGEMENTS

This article was partially supported by the National Institute of Allergy And Infectious Diseases of the National Institutes of Health under Award Numbers R01AI130460 and U24AI171008. We also extend our sincere appreciation to Prof. Xiaoqian Jiang for his support with the GPT experiments in this study.

# ETHICS APPROVAL AND CONSENT TO PARTICIPATE

Not applicable.

## COMPETING INTERESTS

The authors declare that there are no competing interests.

## AUTHOR CONTRIBUTION

Conceptualization, C.T. and Y.L.; methodology, Y.L., and C.T.; software, Y.L. and J.L.; formal analysis, Y.L.; investigation, Y.L.; resources, C.T., H.X., and X.Z.; data curation, D.V. and W.H.; writing—original draft preparation, Y.L.; writing—review and editing, C.T.; visualization, Y.L.; supervision, C.T.; funding acquisition, C.T.; project administration, C.T.. All authors have read and agreed to the published version of the manuscript.

## DATA AVAILABILITY

The datasets generated during and/or analyzed during the current study are available from the corresponding author upon reasonable request.

## ABBREVIATIONS

| | |
|---|---|
| AE | Adverse event |
| AEFI | Adverse events following immunization |
| BERT | Bidirectional Encoder Representations from Transformers |
| BioBERT | Bidirectional Encoder Representations from Transformers for Biomedical Text Mining |
| CDC | Centers for Disease Control and Prevention |
| CLAMP | Clinical Language Annotation, Modeling, and Processing |
| CRF | Conditional Random Field |
| CSV | Comma-Separated-Value |
| FDA | Food and Drug Administration |
| GPT | Generative pre-trained transformer |
| LLM | Large language model |
| NER | Named entity recognition |
| NLP | Natural language processing |
| PT | Preferred Term |
| RNN | Recurrent neural network |
| VAERS | Vaccine Adverse Event Reporting System |

# SUPPLEMENTARY

Supplementary Table S1 Prompts for large language models

| | Model | Prompt Style | Prompt |
|---|---|---|---|
| GPT | Pretrained GPT 2 | Merged | "Please extract all names of dose, vaccine, and adverse event from this note, and put them in a list: {note}" |
| | Fine-tuned GPT 2 | Split | "question": "Please extract all the names of vaccine from the following note"<br>"context": {note} |
| | Pretrained GPT 3.5 | Merged | "role": "system", "content": "Assistant is a large language model trained by OpenAI.",<br>"role": "user", "content": "Please extract all names of vaccine, dose, and adverse event from the following note, and put them in a list:{note}" |
| | Fine-tuned GPT 3.5 | Split | "role": "system", "content": "You are an assistant that is good at named entity recognition.",<br>"role": "user", "content": "Please only extract all |

| | | | *{entity name}* in the following note. Please output the entity directly. Do not contain other information: {note} |
| | Pretrained GPT 4 | Merged | "role": "system", "content": "Assistant is a large language model trained by OpenAI.", "role": "user", "content": "Please extract all names of vaccine, dose, and adverse event from the following note, and put them in a list:{note}" |
| Llama 2 | Pretrained Llama 2 7b | Split | Please extract all the names of {entity name} from the following note:{note} |
| | Fine-tuned Llama 2 7b | Split | |
| | Pretrained Llama 2 13b | Split | |
| | Fine-tuned Llama 2 13b | Split | |

Supplementary Table S2 Parameters for GPT models

| Model | Version | Temperature | Maximum output token |
|---|---|---|---|
| Pretrained GPT 2 | gpt2-large | 1.0 | 1,024 |
| Fine-tuned GPT 2 | gpt2-large | 1.0 | 30 |
| Pretrained GPT 3.5 | gpt-3.5-turbo-16k | 1.0 | 4,096 |
| Fine-tuned GPT 3.5 | gpt-35-turbo-0125 | 0.3 | 4,096 |
| Pretrained GPT 4 | gpt-4-32k | 1.0 | 4,096 |

Supplementary Table S3 Parameters for Llama 2 models

| | Llama 2 7b | Llama 2 13b |
|---|---|---|
| Architectures | LlamaForCausalLM | LlamaForCausalLM |
| Hidden Activation Function | silu | silu |
| Hidden Size | 4,096 | 5,120 |
| Intermediate Size | 11,008 | 13,824 |
| Maximum Position Embeddings | 2,048 | 2,048 |
| Number of Attention Heads | 32 | 40 |

| Number of Hidden Layers | 32 | 40 |
| --- | --- | --- |
| Number of Key Value Heads | 32 | 40 |
| Pretraining TP | 1 | 1 |
| RMS Norm Epsilon | 1e-06 | 1e-05 |
| Tie Word Embeddings | false | false |
| Torch Data Type | float16 | float16 |
| Transformers Version | 4.31.0 | 4.31.0 |
| Use Cache | true | true |
| Vocabulary Size | 32,000 | 32,000 |